\pdfoutput=1

\documentclass[11pt]{article}

\usepackage{naacl2021}
\usepackage{xspace}
\usepackage{adjustbox}
\definecolor{amber}{rgb}{1.0, 0.49, 0.0}
\definecolor{bronze}{rgb}{0.8, 0.5, 0.2}
\definecolor{burntorange}{rgb}{0.9, 0.33, 0.0}
\usepackage{stackengine}

\newcommand{\blue}[1]{\textcolor{blue}{#1}\color{black}\xspace}

\newcommand{\orange}[1]{\textcolor{burntorange}{#1}\color{black}\xspace}

\usepackage{times}
\usepackage{latexsym}
\usepackage{times}
\usepackage{latexsym}
\usepackage{microtype}
\usepackage{multirow, multicol}
\usepackage{cite}
\usepackage{url}

\usepackage{tabularx}
\usepackage{booktabs}
\usepackage{multirow}
\usepackage{microtype}
\usepackage{times}
\usepackage{latexsym}
\usepackage{amsmath}
\usepackage{amssymb}
\usepackage{graphicx}  
\usepackage{lipsum}
\usepackage{multirow}
\usepackage{tabularx}
\usepackage{arydshln}
\usepackage{afterpage}
\usepackage{subcaption}
\usepackage{float}
\usepackage{bbm}
\usepackage{booktabs}
\usepackage{algorithm,algorithmicx,algpseudocode}

\algnewcommand\algorithmicinput{\textbf{ Parameters:}}
\algnewcommand\INPUT{\item[\algorithmicinput]}

\newcommand{\fsc}[1]{\textsc{#1}}

\newcommand{\bF}{\boldsymbol {f}}
\newcommand{\bW}{\boldsymbol {w}}

\newcommand{\pT}{\prod_{t=1}^{T}}
\newcommand{\pC}{\prod_{m=1}^{M}}
\newcommand{\fval}{I_m}
\newcommand{\bGammaM}{\boldsymbol{\gamma}_m}
\newcommand{\Fval}{\boldsymbol{I}}

\newcommand\norm[1]{\left\lVert#1\right\rVert}

\usepackage{todonotes}

\usepackage{cleveref}
\crefname{section}{s}{ss}
\crefname{section}{s}{ss}
\crefname{table}{Table}{}
\crefname{figure}{Fig.}{}
\crefname{algorithm}{Alg.}{}
\crefname{ALC@unique}{Line}{Lines}
\crefname{equation}{Eq.}{}
\crefname{appendix}{Appendix}{}
\crefformat{section}{\S#2#1#3} 
\usepackage[T1]{fontenc}

\usepackage[utf8]{inputenc}

\usepackage{microtype}

\title{Event Representation with Sequential, Semi-Supervised Discrete Variables}

\author{Mehdi Rezaee \\
  Department of Computer Science\\
  University of Maryland Baltimore County\\
  Baltimore, MD 21250 USA \\
  \texttt{rezaee1@umbc.edu} \\
   \And
  Francis Ferraro \\
  Department of Computer Science\\
  University of Maryland Baltimore County\\
  Baltimore, MD 21250 USA \\
  \texttt{ferraro@umbc.edu} \\}

\begin{document}
\maketitle
\begin{abstract}
Within the context of event modeling and understanding, we propose a new method for neural sequence modeling that takes partially-observed sequences of discrete, external knowledge into account. %
We construct a sequential neural variational autoencoder, which uses Gumbel-Softmax reparametrization within a carefully defined encoder, to allow for successful backpropagation during training. %
The core idea is to allow semi-supervised external discrete knowledge to \textit{guide}, but not restrict, the variational latent parameters during training. %
Our experiments indicate that our approach not only outperforms multiple baselines and the state-of-the-art in narrative script induction, but also converges more quickly.
\end{abstract}
\section{Introduction}
\label{sec:Introduction}
Event scripts are a classic way of summarizing events, participants, and other relevant information as a way of analyzing complex situations~\citep{schank1977scripts}. %
To learn these scripts we must be able to group similar-events together, learn common patterns/sequences of events, and learn to represent an event's arguments~\citep{minsky1974}. 
While continuous embeddings can be learned for events and their arguments~\citep{ferraro2017framespr,weber2018event}, the direct inclusion of more structured, discrete knowledge is helpful in learning event representations~\citep{ferraro2016unified}. %
Obtaining fully accurate structured knowledge can be difficult, so when the external knowledge is neither sufficiently reliable nor present, a natural question arises: how can our models use the knowledge that \textit{is} present? %

Generative probabilistic models provide a framework for doing so: external knowledge is a random variable, which can be observed or latent, and the data/observations are generated (explained) from it. %
Knowledge that is discrete, sequential, or both---such as for script learning---complicates the development of neural generative models. %

In this paper, we provide a successful approach for incorporating partially-observed, discrete, sequential external knowledge in a neural generative model. %
We specifically examine \textit{event sequence modeling} augmented by \textit{semantic frames}. %
Frames~\citep[i.a.]{minsky1974} are a semantic representation designed to capture the common and general knowledge we have about events, situations, and things. %
They have been effective in providing crucial information for modeling and understanding the meaning of events~\citep{peng-2016-semlm,ferraro2017framespr,padia-2018-fever,zhang2020semanticsbert}. 
Though we focus on semantic frames as our source of external knowledge, we argue this work is applicable to other similar types of
knowledge. %

\begin{figure}[t]
    \centering
    \includegraphics[width=0.99\columnwidth]{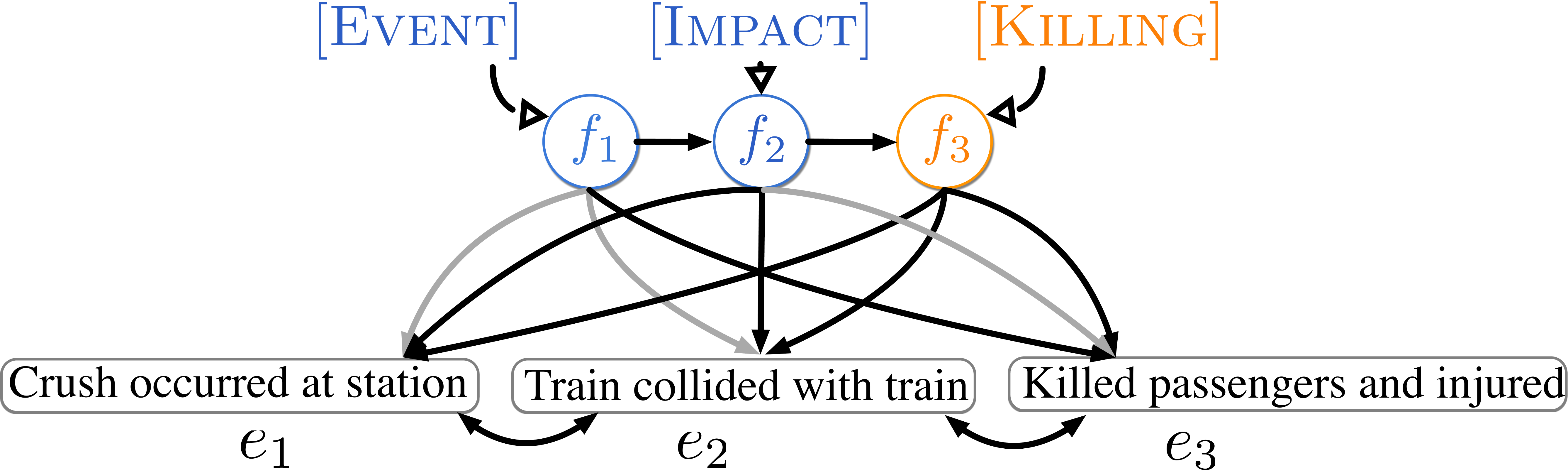}
    \caption{%
    An overview of event modeling, where the observed events (black text) are generated via a sequence of semi-observed random variables. %
    In this case, the random variables are directed to take on the meaning of semantic frames that can be helpful to explain the events. %
    Unobserved frames are in orange and observed frames are in blue. %
    Some connections are more important than others, indicated by the weighted arrows. %
    }
    \label{fig:toy_example}
    \vspace{-4mm}
  \end{figure}

We examine the problem of modeling observed \textit{event tuples} as a partially observed sequence of \textbf{semantic frames}. %
Consider the following three events, preceded by their corresponding bracketed frames, from \cref{fig:toy_example}:

\vspace{0.05in}
\parbox{2.9in}{\textit{\fsc{[{\textbf{Event}}]} crash occurred at station.\\ \fsc{[\textbf{Impact}]} train collided with train.\\
\fsc{[\textbf{Killing}]} killed passengers and injured}.} \\ 
\vspace{0.01in}

\noindent
We can see that even without knowing the $\fsc{[\textbf{Killing}]}$ frame, the $\fsc{[\textbf{Event}]}$ and $\fsc{[\textbf{Impact}]}$ frames can help predict the word \textit{killed} in the third event; the frames summarize the events and can be used as guidance for the latent variables to represent the data.
On the other hand, words like \textit{crash}, \textit{station} and \textit{killed} from the first and third events play a role in predicting [\fsc{\textbf{Impact}}] in the second event. %
Overall, to successfully represent events, beyond capturing the event to event connections, we propose to consider all the information from the frames to events and frames to frames. %

In this work, we study the effect of tying discrete, sequential latent variables to partially-observable, noisy (imperfect) semantic frames. %
Like \citet{weber2018hierarchical}, our semi-supervised model is a bidirectional auto-encoder, with a structured collection of latent variables separating the encoder and decoder, and attention mechanisms on both the encoder and decoder.
Rather than applying vector quantization, we adopt a Gumbel-Softmax~\citep{jang2017categorical} ancestral sampling method to easily switch between the observed frames and latent ones, where we inject the observed frame information on the Gumbel-Softmax parameters before sampling. %
Overall, our contributions are: 
\begin{itemize}
\setlength{\itemsep}{0pt}
    \item We demonstrate how to learn a VAE that contains sequential, discrete, and \textit{partially-observed} latent variables. %
    \item We show that adding partially-observed, external, semantic frame knowledge to  our structured, neural generative model leads to improvements over the current state-of-the-art on recent core event modeling tasks. %
    Our approach leads to faster training convergence. %
    \item We show that our semi-supervised model, though developed as a generative model, can effectively predict the labels that it may not observe. Additionally, we find that our model outperforms a discriminatively trained model with full supervision. %
\end{itemize}
 \begin{figure*}[t]
    \centering
    \includegraphics[width=1.0\textwidth]{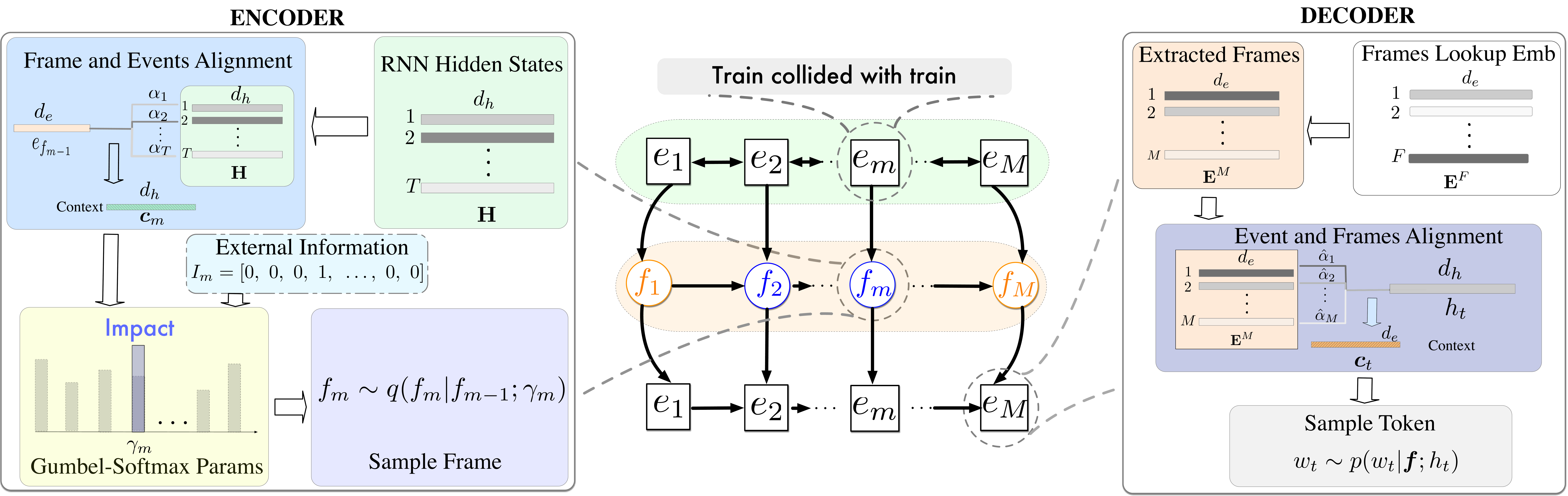}
    \caption{Our encoder (left) and decoder (right).  The orange nodes mean that the frame is latent ($\fval=\boldsymbol{0}$), while the blue nodes indicate observed frames ($\fval$ is one-hot). In the encoder, the RNN hidden vectors are 
    aligned with the frames to predict the next frame. The decoder utilizes the inferred frame information in the reconstruction.
    }
    \label{fig:Encoder-Decoder}
    \vspace{-4mm}
\end{figure*} 

\section{Related Work}
\label{sec:related_work}
Our work builds on both event modeling and latent generative models. In this section, we outline relevant background and related work.

\subsection{Latent Generative Modeling}
\label{sec:background}
Generative latent variable models learn a mapping from the low-dimensional hidden variables $\mathbf{f}$ to the observed data points $\mathbf{x}$, where the hidden representation captures the high-level information to explain the data. Mathematically, the joint probability $p(\mathbf{x},\mathbf{f};\theta)$ factorizes as follows
\begin{align}
    p(\mathbf{x},\mathbf{f};\theta)=
    p(\mathbf{f})
    p(\mathbf{x}\vert \mathbf{f};\theta),
\end{align}
where $\theta$ represents the model parameters. Since in practice maximizing the log-likelihood is intractable, we approximate the posterior by defining $q(\mathbf{f}\vert \mathbf{x};\phi)$ and maximize the ELBO \citep{kingma2013auto} as a surrogate objective:
\begin{align}
    \mathcal{L}
    _{\theta,\phi}
    =
    \mathbb{E}
    _
    {q(\mathbf{f}|\mathbf{x};\phi)}
    \log
    \dfrac
    {p(\mathbf{x},\mathbf{f};\theta)}
    {q(\mathbf{f}\vert \mathbf{x};\phi)}.
\end{align}
In this paper, we are interested in studying a specific case; the input $\mathbf{x}$ is a sequence of $T$ tokens, we have $M$ sequential discrete latent variables $\mathbf{z}=\{z_m\}_{m=1}^{M}$, where each $z_m$ takes $F$ discrete values. While there have been effective proposals for unsupervised optimization of $\theta$ and $\phi$, we focus on learning \textit{partially observed} sequences of these variables. That is, we assume that in the training phase some values are observed while others are latent. We incorporate this partially observed, external knowledge to the $\phi$ parameters to guide the inference. The inferred latent variables later will be used to reconstruct to the tokens.

\citet{kingma2014semi} generalized VAEs to the semi-supervised setup, but they assume that the dataset can be split into observed and unobserved samples and they have defined separate loss functions for each case; in our work, we allow portions of a sequence to be latent. \Citet{teng2020semi} characterized the semi-supervised VAEs via sparse latent variables; see \citet{mousavi2019survey} for an in-depth study of additional sparse models.

Of the approaches that have been developed for handling discrete latent variables in a neural model~\citep[i.a.]{vahdat2018dvae,vahdat2018dvaepp,lorberbom2019direct}, we use the Gumbel-Softmax reparametrization~\citep{jang2017categorical,maddison2016concrete}. %
This approximates a discrete draw with logits $\pi$ as $\mathrm{softmax}(\frac{\pi + g}{\tau})$, where $g$ is a vector of $\mathrm{Gumbel}(0,1)$ draws and $\tau$ is an annealing temperature that allows targeted behavior; its ease, customizability, and efficacy are big advantages. %

\subsection{Event Modeling}
Sequential event modeling, as in this paper, can be viewed as a type of \textit{script} or \textit{schema} induction~\citep{schank1977scripts} via language modeling techniques. %
\Citet{mooney1985learning} provided an early analysis of explanatory schema generating system to process narratives, and \citet{pichotta2016learning} applied an LSTM-based model to predict the event arguments. %
\Citet{modi2016event} proposed a neural network model to predict randomly missing events, while \citet{rudinger-etal-2015-script} showed how neural language modeling can be used for sequential event prediction. %
\Citet{weber2018event} and \Citet{ding2019event} used tensor-based decomposition methods for event representation. 
\Citet{weber-etal-2020-causal} studied causality in event modeling via a latent neural language model. %

Previous work has also examined how to incorporate or learn various forms of semantic representations while modeling events. %
\Citet{cheung2013probabilistic} introduced an HMM-based model to explain event sequences via latent  frames. 
\Citet{materna2012ldaframes}, \Citet{chambers2013event} and \citet{bamman2013personas} provided structured graphical models to learn event models over syntactic dependencies; \citet{ferraro2016unified} unified and generalized these approaches to capture varying levels of semantic forms and representation. %
\Citet{kallmeyer2018coarse} proposed a Bayesian network based on a hierarchical dependency between syntactic dependencies and frames. %
\Citet{Ribeiro2019L2FINESCIDAS} provided an analysis of clustering predicates and their arguments to infer semantic frames. %

Variational autoencoders and attention networks \citep{kingma2013auto,bahdanau2014neural},
allowed \citet{bisk2019benchmarking} to use RNNs with attention to capture the abstract and concrete script representations. %
\Citet{weber2018hierarchical} came up with a recurrent autoencoder model (HAQAE), which used vector-quantization to learn hierarchical dependencies among discrete latent variables and an observed event sequence. %
\citet{kiyomaru2019diversity} suggested generating next events using a conditional VAE-based model. %

In another thread of research, \citet{chen2018variational} utilized labels in conjunction with latent variables, but unlike \citet{weber2018hierarchical}, their model's latent variables are conditionally independent and do not form a hierarchy. %
\Citet{sonderby2016ladder} proposed a sequential latent structure with Gaussian latent variables. %
\Citet{lievin2019towards}, similar to our model structure, provided an analysis of hierarchical relaxed categorical sampling but for the unsupervised settings.


\section{Method}
\label{sec:Method}
Our focus in this paper is modeling sequential event structure. In this section, we describe our variational autoencoder model, and demonstrate how partially-observed external knowledge can be injected into the learning process. %
We provide an overview of our joint model in \cref{sec:model_setup} and \cref{fig:Encoder-Decoder}. %
Our model operates on sequences of events: it consists of an encoder (\cref{sec:Encoder}) that encodes the sequence of events as a new sequence of frames (higher-level, more abstract representations), and a decoder (\cref{sec:Decoder}) that learns how to reconstruct the original sequence of events from the representation provided by the encoder. %
During training (\cref{sec:Training_Process}), the model can make use of partially-observed sequential knowledge to enrich the representations produced by the encoder. %
In \cref{sec:model_novelties} we summarize the novel aspects of our model.

\subsection{Model Setup}
\label{sec:model_setup}
We define each document as a sequence of $M$ events. %
In keeping with previous work on event representation, each event is represented as a lexicalized 4-tuple:
the core event predicate (verb), two main arguments (subject and object), and event modifier (if applicable)~\citep{pichotta2016learning,weber2018hierarchical}. %
For simplicity, we can write each document as a sequence of $T$ words $\mathbf{w}=\{w_{t}\}_{t=1}^{T}$, where $T=4M$ and each $w_t$ is from a vocabulary of size $V$.\footnote{\Cref{tab:notations} in the Appendix provides all the notations.} \Cref{fig:toy_example} gives an example of 3 events: during learning (but not testing) our model would have access to some, but not all, frames to lightly guide training (in this case, the first two frames but not the third).

While lexically rich, this 4-tuple representation is limited in the knowledge that can be directly encoded. %
Therefore, our model assumes that each document $\mathbf{w}$ can be explained jointly with a collection of $M$ random variables $f_m$: $\boldsymbol{f}=\{f_{m}\}_{m=1}^{M}$ . %
The joint probability for our model factorizes as
\begin{equation}
  p(\bW,\bF)
  =
  \pT{p(w_t\vert \bF,w_{<t})}\pC p(f_m\vert f_{m-1}).
\label{eq:joint_model}
\end{equation}

\noindent For event modeling, each $f_m$ represents a semantic frame. %
We assume there are $F$ unique frames and let $f_m$ be a discrete variable indicating which frame, if any, was triggered by event $m$.\footnote{Our model is theoretically adaptable to making use of multiple frames, though we assume each event triggers at most one frame.}

In the general case, $\mathbf{f}$ is completely unobserved, and so inference for this model requires marginalizing over $\mathbf{f}$: when $F \gg 1$, optimizing the likelihood is intractable. %
We follow amortized variational inference \citep{kingma2013auto} as an alternative approach and use an ancestral sampling technique to compute it. %
We define $q(\bF \vert \bW)$ as the variational distribution over $\bF$, which can be thought of as stochastically encoding $\bW$ as $\bF$.

Our method is semi-supervised, so we follow \citet{kingma2014semi}, \citet{chen2018variational} and \citet{ye2020variational} and optimize a weighted variant of the evidence lower bound (ELBO),  %
\begin{equation}
\begin{aligned}
    \mathcal{L} = &
    \overbrace{\mathbb{E}_{q(\bF\vert \bW)}
    \log
    {
        p(\bW\vert \bF)
    }}^{\text{Reconstruction term}}
    +
    \overbrace{\alpha_q\mathbb{E}_{q(\bF\vert \bW)}
    \log
    \dfrac
    {
        p(\bF)
    }
    {
        q(\bF\vert \bW)
    }}^{\text{KL term}},
    \\
     & + \overbrace{\alpha_c \mathcal{L}_c( q(\bF\vert \bW)),
    }^{\text{Supervised classification term}}
\end{aligned}
\label{eq:ELBO}
\end{equation}
where $\mathcal{L}_c( q(\bF\vert \bW))$ is a classification objective that encourages $q$ to predict the frames that actually were observed, and $\alpha_q$ and $\alpha_c$ are empirically-set to give different weight to the KL vs. classification terms. We define $\mathcal{L}_c$ in \cref{sec:Training_Process}. %
The reconstruction term learns to generate the observed events $\bW$ across all valid encodings $\bF$, while the KL term uses the prior $p(\bF)$ to regularize $q$. %

Optimizing \cref{eq:ELBO} is in general intractable, so we sample $S$ chains of variables $\bF^{(1)}, \ldots, \bF^{(S)}$ from $q(\bF \vert \bW)$ and approximate \cref{eq:ELBO} as
\begin{equation}
\begin{aligned}
    \mathcal{L} \approx &
    \frac{1}{S} \sum_{s} 
    \left[
    \log
    {
        p(\bW\vert \bF^{(s)})
    } +
    \alpha_q\log
    {
        \frac{p(\bF^{(s)})}{q(\bF^{(s)} | \bW)}
    } + \right.\\
    & \left.\alpha_c \mathcal{L}_c( q(\bF^{(s)}\vert \bW))\right].
    \label{eqn:approx-elbo}
    \end{aligned}
\end{equation}

As our model is designed to allow the injection of external knowledge $\Fval$, we define the variational distribution as $q(\bF | w; I)$. %
In our experiments, $\fval$ is a binary vector encoding which (if any) frame is observed for event $m$.\footnote{If a frame is observed, then $\fval$ is a one-hot vector where the index of the observed frame is 1. Otherwise  $\fval=\overrightarrow{\boldsymbol{0}}$. %
} %
For example in \cref{fig:toy_example}, we have $I_1=1$ and $I_3=0$. %
We define
\begin{equation}
  q(\bF\vert \bW;\Fval)
  =  \pC q(f_{m}\vert f_{m-1},\fval,\bW).
  \label{eqn:variational-distribution}
\end{equation}
We broadly refer to \cref{eqn:variational-distribution} as our \textbf{encoder}; we detail this in \cref{sec:Encoder}. In \cref{sec:Decoder} we describe how we compute the reconstruction term, and in \cref{sec:Training_Process} we provide our  semi-supervised training procedure. %

\subsection{Encoder} 
\label{sec:Encoder}
The reconstruction term relies on the frame samples given by the encoder. %
As discussed above though, we must be able to draw chains of variables $\bF$, by iteratively sampling $f_{m}\sim q(\cdot\vert f_{m-1},\bW; \Fval)$, in a way that allows the external knowledge $\Fval$ to \textit{guide}, but not restrict, $\bF$. %
This is a deceptively difficult task, as the encoder must be able to take the external knowledge into account in a way that neither prevents nor harms back-propagation and learning. %
We solve this problem by learning to compute a good representation $\bGammaM$ for each event, and sampling the current frame ${f_m}$ from a Gumbel-Softmax distribution~\citep{jang2017categorical} parametrized by $\bGammaM$. %

\begin{algorithm}[t]
    \begin{algorithmic}[1]    
    \Require 
    previous frame $f_{m-1}$,
    \Comment $f_{m-1}\in \mathbb{R}^F$
    
    \noindent \ \ \ \ \ \ current frame observation ($\fval$),
    
    \noindent \ \ \ \ \ \ encoder GRU hidden states $\mathbf{H}\in \mathbb{R}^{T\times d_h}$.
    
    \noindent\INPUT $W_{\text{in}}\in \mathbb{R}^{d_h\times d_e}$, $W_{\text{out}}\in \mathbb{R}^{F\times d_h}$,
    
    \noindent \ \ \ \ \ \ frames embeddings $\boldsymbol{E}^F \in \mathbb{R}^{F\times d_e}$
    \Ensure  ${f_m}$, $\bGammaM$
    \State  $e_{f_{m-1}}={f_{m-1}}^{\top}\boldsymbol{E}^{F}$\label{alg:Encoder_frame_embedding}
    \State ${\boldsymbol{\alpha}}\gets \mathrm{Softmax}(\mathbf{H}W_{\text{in}}e_{f_{m-1}}^{\top})$ \Comment{Attn. Scores}
    \label{alg:Encoder_alpha}
    \State ${\boldsymbol{c}_m} 
    \gets \boldsymbol{H}^{\top}\boldsymbol{\alpha}$  
    \label{alg:Encoder_context}
    \Comment{Context Vector}
    \State {$\bGammaM' \gets W_{\text{out}}\big(\tanh(W_{\text{in}}e_{f_{m-1}})+\tanh(\boldsymbol{c}_m)\big)$ }
     \label{alg:Encoder_gamma_raw}
    \State $\bGammaM \gets \bGammaM +  \norm{\bGammaM}{\fval}$ \Comment{Observation}
   \label{alg:Encoder_inf_added}        
    \State $q(f_m\vert f_{m-1})\gets \mathrm{GumbelSoftmax}(\bGammaM)$

    \State ${f_m} \sim q(f_m\vert f_{m-1})$
    \label{alg:Encoder_fm_sample}    
    \Comment{${f_m}\in \mathbb{R}^F$}
    \end{algorithmic}
    \caption{Encoder: The following algorithm shows how we compute the next frame $f_m$ given the previous frame $f_{m-1}$. %
    We compute and return a hidden frame representation $\bGammaM$, and $f_m$ via a continuous Gumbel-Softmax reparametrization. %
    \label{alg:Encoder}
    }
  \end{algorithm}

\Cref{alg:Encoder} gives a detailed description of our encoder. %
We first run our event sequence through a recurrent network (like an RNN or bi-LSTM); if $\bW$ is $T$ tokens long, this produces $T$ hidden representations, each of size $d_h$. Let this collection be $\mathbf{H}\in \mathbb{R}^{T\times d_h}$. %
Our encoder proceeds iteratively over each of the $M$ events as follows: given the previous sampled frame ${f_{m-1}}$, the encoder first computes a weighted embedding $e_{f_{m-1}}$ of this previous frame (\cref{alg:Encoder_frame_embedding}). Next, it calculates the similarity between $e_{f_{m-1}}$ and RNN hidden representations and all recurrent hidden states $\mathbf{H}$ (\cref{alg:Encoder_alpha}). %
After deriving the attention scores, the weighted average of hidden states ($\boldsymbol{c}_m$) summarizes the role of tokens in influencing the frame $f_m$ for the $m$\textsuperscript{th} event (\cref{alg:Encoder_context}). %
We then combine the previous frame embedding $e_{f_{m-1}}$ and the current context vector $\boldsymbol{c}_m$ to obtain a representation $\bGammaM'$ for the $m$\textsuperscript{th} event (\cref{alg:Encoder_gamma_raw}). 

While $\bGammaM'$ may be an appropriate representation if no external knowledge is provided, our encoder needs to be able to inject any provided external knowledge $\fval$. %
Our model defines a chain of variables---some of which may be observed and some of which may be latent---so care must be taken to preserve the gradient flow within the network. %
We note that an initial strategy of solely using ${\fval}$ instead of ${f_m}$ (whenever $\fval$ is provided) is not sufficient to ensure gradient flow. %
Instead, we incorporate the observed information given by $\fval$ by adding this information to the output of the encoder logits before drawing ${f_m}$ samples (\cref{alg:Encoder_inf_added}). %
This remedy motivates the encoder to \textit{softly} increase the importance of the observed frames during the training. %
Finally, we draw $f_m$ from the Gumbel-Softmax distribution (\cref{alg:Encoder_fm_sample}). %

For example, in \cref{fig:toy_example}, when the model knows that \fsc{[\textbf{Impact}]} is triggered, it increases the value of \fsc{[\textbf{Impact}]} in $\bGammaM$ to encourage \fsc{[\textbf{Impact}]} to be sampled, but it does not prevent other frames from being sampled.  On the other hand, when a frame is not observed in training, such as for the third event (\fsc{[\textbf{Killing}]}), $\bGammaM$ is \textit{not} adjusted.

Since each draw $f_m$ from a Gumbel-Softmax is a simplex vector, given learnable frame embeddings $\boldsymbol{E}^{F}$, we can obtain an aggregate frame representation $e_m$ by calculating $e_m={f_m}^{\top}{\boldsymbol{E}^{F}}$. %
This can be thought of as roughly extracting row $m$ from $\boldsymbol{E}^{F}$ for low entropy draws $f_m$, and using many frames in the representation for high entropy draws. %
Via the temperature hyperparameter, the Gumbel-Softmax allows us to control the entropy. %

\subsection{Decoder} 
\label{sec:Decoder}

Our decoder (\cref{alg:Decoder}) must be able to reconstruct the input event token sequence from the frame representations $\bF = (f_1, \ldots, f_M)$ computed by the encoder. %
In contrast to the encoder, the decoder is relatively simple: we use an auto-regressive (left-to-right) GRU to produce hidden representations $z_t$ for each token we need to reconstruct, but we enrich that representation via an attention mechanism over $\bF$. %
Specifically, we use both $\bF$ and the same learned frame embeddings $\mathbf{E}^F$ from the encoder to compute inferred, contextualized frame embeddings as $\boldsymbol{E}^M =\bF\mathbf{E}^F \in \mathbb{R}^{M\times d_e}$. %
For each output token (time step $t$), we align the decoder GRU hidden state $h_t$ with the rows of $\boldsymbol{E}^M$ (\cref{alg:Decoder_align}). %
After calculating the scores for each frame embedding, we obtain the output context vector $\boldsymbol{c}_t$ (\cref{alg:Decoder_context}), which is used in conjunction with the hidden state of the decoder $z_{t}$ to generate the $w_t$ token (\cref{alg:Decoder_recons}). %
In \cref{fig:toy_example}, the collection of all the three frames and the
tokens from the first event will be used to predict the \textit{Train} token from the second event.


\subsection{Training Process} 
\label{sec:Training_Process}
\begin{algorithm}[t]
  \begin{algorithmic}[1]    
  \Require  
  $\boldsymbol{E}^M \in \mathbb{R}^{M\times d_e}$ (computed as $\bF\mathbf{E}^F$)

  \noindent \ \ \ \ \ \ decoder's current hidden state $z_{t}\in \mathbb{R}^{d_h}$
  
  \noindent \INPUT $\hat{W}_{\text{in}}\in \mathbb{R}^{d_e\times d_h}$, $\hat{W}_{\text{out}}\in \mathbb{R}^{V\times d_e}$, $\mathbf{E}^F$
  
  \Ensure  $w_t$
  \State $  {\boldsymbol{\hat{\alpha}}}\gets \mathrm{Softmax}({\boldsymbol{E}^M}\hat{W}_{\text{in}}z_t)$
  \label{alg:Decoder_align}
  \Comment{Attn. Scores}
  \State ${\boldsymbol{c}_t} \gets
  {\boldsymbol{E}^M}^{\top}{\boldsymbol{\hat{\alpha}}}$
  \label{alg:Decoder_context}
  \Comment{Context Vector}
  \State {$g\gets \hat{W}_{\text{out}}\big(\tanh(\hat{W}_{\text{in}}z_t)+\tanh(\boldsymbol{c}_t)\big)$ }
  \label{alg:Decoder_addition}
  \State
  $p(w_t\vert \bF;z_t)\propto \exp(g)$    
  \State
  $w_t \sim p(w_t\vert \bF;z_t)$
\label{alg:Decoder_recons}
  \end{algorithmic}
\caption{Decoder: To (re)generate each token in the event sequence, we compute an attention $\boldsymbol{\hat\alpha}$ over the sequence of frame random variables $\bF$ (from \cref{alg:Encoder}). %
This attention weights each frame's contribution to generating the current word. %
  \label{alg:Decoder}
  }
\end{algorithm}
We now analyze the different terms in \cref{eqn:approx-elbo}. %
In our experiments, we have set the number of samples $S$ to be 1. %
From \cref{eqn:variational-distribution}, and using the sampled sequence of frames $f_1,f_2, \ldots f_M$ from our encoder, we approximate the reconstruction term as
$
    \mathcal{L}_w = \sum_t \log
    {
        p(w_t \vert f_1,f_2, \ldots f_M; z_t)
    },
$
where $z_t$ is the decoder GRU's hidden representation after having reconstructed the previous $t-1$ words in the event sequence.

Looking at the KL-term, we define the \textit{prior} frame-to-frame distribution as $p(f_m\vert f_{m-1})=1/F$, and let the variational distribution capture the dependency between the frames. %
A similar type of strategy has been exploited successfully by \citet{chen2018variational} to make computation simpler. %
We see the computational benefits of a uniform prior: splitting the KL term into
$
    \mathbb{E}_{q(\bF\vert \bW)}
    \log
    {
        p(\bF)
    }
    -
    \mathbb{E}_{q(\bF\vert \bW)}
    \log
    {
        q(\bF\vert \bW)
    }
$ 
allows us to neglect the first term. %
For the second term, we normalize the Gumbel-Softmax logits $\bGammaM$, i.e., $\bGammaM = \mathrm{Softmax}(\bGammaM)$, and compute
\begin{align*}
    \mathcal{L}_q 
    &=
    -\mathbb{E}_{q(\bF\vert \bW)}
    \log
    {
        q(\bF\vert \bW)
    }    
    \approx
-\sum_{m}
\bGammaM^{\top} \log \bGammaM.
\end{align*}
$\mathcal{L}_q $ encourages the entropy of the variational distribution to be high which makes it hard for the encoder
to distinguish the true frames from the wrong ones. %
We add a \textit{fixed and constant} regularization coefficient $\alpha_q$ to decrease the effect of this term~\citep{bowman2015generating}. %
We define the classification loss as
$
    \mathcal{L}_c
    =
-\sum_{\fval>0}
    {\fval}^{\top} \log \bGammaM,
    $
to encourage $q$ to be good at predicting any frames that were actually observed.
We weight $\mathcal{L}_c$ by a fixed coefficient $\alpha_c$. %
Summing these losses together, we arrive at our objective function:
\begin{equation}
    \mathcal{L} = \mathcal{L}_w + \alpha_q \mathcal{L}_q + \alpha_c\mathcal{L}_c.
\end{equation}
\subsection{Relation to prior event modeling}
\label{sec:model_novelties}
A number of efforts have leveraged frame induction for event modeling~\citep{cheung2013probabilistic,chambers2013event,ferraro2016unified,kallmeyer2018coarse,Ribeiro2019L2FINESCIDAS}. 
These methods are restricted to explicit connections between events and their corresponding frames; they do not capture all the possible connections between the observed events and frames. %
\Citet{weber2018hierarchical} proposed a hierarchical unsupervised attention structure (HAQAE) that corrects for this. %
HAQAE uses vector quantization~\citep{van2017neural} to capture sequential event structure via tree-based latent variables. 

Our model is related to HAQAE~\citep{weber2018hierarchical}, though with important differences. %
While HAQAE relies on unsupervised deterministic inference, we aim to incorporate the frame information in a softer, guided fashion. %
The core differences are: our model supports partially-observed frame sequences (i.e., semi-supervised learning); the linear-chain connection among the event variables in our model is simpler than the tree-based structure in HAQAE; while both works use attention mechanisms in the encoder and decoder, our attention mechanism is based on addition rather than concatenation; and our handling of latent discrete variables is based on the Gumbel-Softmax reparametrization, rather than vector quantization. We discuss these differences further in \cref{app:haqae_review}.
\section{Experimental Results}
\label{sec:Experimental Result}
We test the performance of our model on a portion of the Concretely Annotated Wikipedia dataset~\citep{ferraro-2014-annotated}, which is a dump of English Wikipedia that has been annotated with the outputs of more than a dozen NLP analytics; we use this as it has readily-available FrameNet annotations provided via SemaFor~\citep{das2014frame}. %
Our training data has 457k documents, our validation set has 16k documents, and our test set has 21k documents. %
More than 99\% of the frames  
are concentrated in the first 500 most common frames, so we set $F=500$. %
Nearly 15\% of the events did not have
any frame, many of which were due to auxiliary/modal verb structures; as a result, we did not include them. %
For all the experiments, the vocabulary size $(V)$ is set as $40k$ and the number of events $(M)$ is 5; this is to maintain comparability with HAQAE. %
For the documents that had more than 5 events, we extracted the first 5 events that had frames. %
For both the validation and test datasets, we have set $\fval=0$ for all the events; frames are only observed during training.

\begin{table*}
\begin{subtable}[t]{.41\textwidth}
    \centering
    \resizebox{\columnwidth}{!}{
    \begin{tabular}{l|c|c|c}
    
    \specialrule{.1em}{.1em}{0.01em}
    \multicolumn{1}{c|}{\multirow{3}{*}{Model}} & \multicolumn{1}{|c|}{\multirow{3}{*}
    {$\epsilon$}
    } & \multicolumn{2}{|c}{PPL} \\
    \cline{3-4} 
    \multicolumn{1}{c|}{} & \multicolumn{1}{c|}{} & Valid & Test  \\ \hline
    RNNLM & - 
    & 61.34 $\pm{ 2.05 }$
    &  61.80 $\pm{ 4.81     }$
    \\ 
     RNNLM+ROLE & - 
     & 66.07 $\pm{ 0.40     }$
     &  60.99 $\pm{ 2.11     }$
     \\ 
    HAQAE & - 
    & 24.39 $\pm{ 0.46 }$
    & 21.38 $\pm{ 0.25 }$
    \\ \cdashline{1-4}
    Ours & 0.0 
    & 41.18 $\pm{ 0.69 }$
    &  36.28 $\pm{ 0.74 }$
    \\ 
    Ours & 0.2 
    & 38.52 $\pm{ 0.83 }$
    & 33.31 $\pm{ 0.63}$
    \\ 
    Ours & 0.4 
    & 37.79 $\pm{0.52}$
    & 33.12 $\pm{0.54}$
    \\ 
    Ours & 0.5 
    & 35.84 $\pm{0.66}$
    &  31.11 $\pm{0.85}$
    \\   
    Ours & 0.7 
    & 24.20 $\pm{1.07 }$
    &  21.19 $\pm{0.76}$
    \\   
    Ours & 0.8 
    & 23.68 $\pm{0.75}$
    &  20.77 $\pm{0.73}$
    \\   
    Ours & 0.9 
    & \textbf{22.52} $\pm{\mathbf{0.62}}$
    &  \textbf{19.84} $\pm{\mathbf{0.52}}$
    \\ 
    \specialrule{.1em}{.1em}{0.005em}
\end{tabular}
}
    \caption{Validation and test per-word perplexities (lower is better).  %
    We always outperform RNNLM and RNNLM+ROLE, and outperform HAQAE when automatically extracted frames are sufficiently available during training ($\epsilon \in \{0.7, 0.8, 0.9\}$). %
    }
    \label{tab:Perplexity}
\end{subtable}
\begin{subtable}[t]{.58\textwidth}
    \centering
    \resizebox{.98\columnwidth}{!}{
    \begin{tabular}{l|c|c|c|c|ccc}
    \specialrule{.1em}{.05em}{.05em} 
    \multicolumn{1}{c|}{\multirow{3}{*}{Model}} & \multicolumn{1}{|c|}{\multirow{3}{*}{$ \epsilon $}}  & \multicolumn{4}{|c}{Inv Narr Cloze} \\ \cline{3-6} 
    \multicolumn{1}{c|}{} & \multicolumn{1}{|c|}{}  & \multicolumn{2}{|c|}{Wiki} & \multicolumn{2}{|c}{NYT} \\ \cline{3-8} 
    \multicolumn{1}{c|}{} & \multicolumn{1}{c|}{} & Valid & Test & Valid & Test \\ \hline
     RNNLM       & -  
     & 20.33  $\pm 0.56$     
     &  21.37 $\pm 0.98 $
     & 18.11 $\pm 0.41  $
     & 17.86 $\pm 0.80  $
     \\ 
     RNNLM+ROLE  & -  
     & 19.57 $\pm  0.68    $
     & 19.69 $\pm 0.97  $
     & 17.56 $\pm 0.10  $
     & 17.95 $\pm 0.25   $
     \\ 
     HAQAE       & -  
     & 29.18 $\pm 1.40   $
     &  24.88 $\pm 1.35    $
     & 20.5 $\pm 1.31  $
     & 22.11 $\pm 0.49   $
     \\ \cdashline{1-6}
     Ours & 0.0  
     &  43.80 $\pm 2.93  $
     & 45.75 $\pm 3.47     $
     & 29.40 $\pm 1.17  $
     & 28.63 $\pm 0.37   $
     \\ 
     Ours & 0.2  
     &  45.78 $\pm 1.53  $
     &  44.38 $\pm 2.10$
     & 	29.50 $\pm 1.13  $
     &  29.30 $\pm 1.45 $
     \\ 
     Ours & 0.4  
     &  \textbf{47.65} $\mathbf{\pm 3.40}$
     & 	\textbf{47.88} $\mathbf{\pm 3.59}$
     &  \textbf{30.01} $\mathbf{\pm 1.27}$
     &  \textbf{30.61} $\mathbf{\pm 0.37}$
     \\ 
     Ours & 0.5  
     &  42.38 $\pm 	2.41  $
     &  40.18 $\pm 0.90    $
     &  29.36 $\pm 	1.58  $
     &  29.95 $\pm 0.97  $
     \\ 
     Ours & 0.7  
     & 38.40 $\pm 1.20   $
     & 39.08 $\pm 1.55  $
     & 29.15 $\pm 0.95$
     & 30.13 $\pm 0.66  $
     \\ 
     Ours & 0.8  
     & 	39.48 $\pm 3.02   $
     & 	38.96 $\pm 2.75     $
     &  29.50 $\pm 0.30 $
     &  30.33 $\pm 0.81  $
     \\ 
     Ours & 0.9  
     &  35.61 $\pm 0.62  $
     &  35.56 $\pm 1.70    $
     &  28.41 $\pm 0.29 $
     &  29.01 $\pm 0.84 $
     \\
    \specialrule{.1em}{.05em}{.05em} 
    \end{tabular}
    }
    \caption{Inverse Narrative Cloze scores (higher is better), averaged across 3 runs, with standard deviation reported. Some frame observation ($\epsilon=0.4$) is most effective across Wikipedia and NYT, though we outperform our baselines, including the SOTA, at any level of frame observation. For the NYT dataset, we first trained the model on the Wikipedia dataset and then did the tests on the NYT valid and test inverse narrative cloze datasets.
    } 
    \label{tab:invNarrClozeStats}
\end{subtable}
\caption{ %
Validation and test results for  per-word perplexity (\cref{tab:Perplexity}: lower is better) and inverse narrative cloze accuracy (\cref{tab:invNarrClozeStats}: higher is better).  %
Recall that $\epsilon$ is the (average) percent of frames observed during \textit{training} though during evaluation \textit{no} frames are observed. %
}
\label{tab:ppl_inv}
\end{table*}

Documents are fed to the model as a sequence of events with verb, subj, object and modifier elements. %
The events are separated with a special separating \textsc{<tup>} token and the missing elements are represented with a special \textsc{None} token. %
In order to facilitate semi-supervised training and examine the impact of frame knowledge, we introduce a user-set value $\epsilon$: in each document, for event $m$, the true value of the frame is preserved in $\fval$ with probability $\epsilon$, while
with probability $1-\epsilon$ we set $\fval=0$. %
This $\epsilon$ is set and fixed prior to each experiment. %
For all the experiments we set $\alpha_q$ and $\alpha_c$ as 0.1, found empirically on the validation data.

\paragraph{Setup}
We represent words by their pretrained Glove 300 embeddings and used gradient clipping at 5.0 to prevent exploding gradients. %
We use a two layer of bi-directional GRU for the encoder, and a two layer uni-directional GRU for the decoder (with a hidden dimension of 512 for both). %
See \cref{app:setup_details} for additional computational details.\footnote{\url{https://github.com/mmrezaee/SSDVAE}} %
\paragraph{Baselines}
In our experiments, we compare our proposed methods against the following methods:
\begin{itemize}
\setlength{\itemsep}{0mm}
        \item \textbf{RNNLM}: We report the performance of a sequence to sequence language model with the same structure used in our own model. 
        A Bi-directional GRU cell with two layers, hidden dimension of 512, gradient clipping at 5 and Glove 300 embeddings to represent words. 
        \item \textbf{RNNLM+ROLE}~
        \citep{pichotta2016learning}:
        This model has the same structure as RNNLM, but the role for each token (verb, subject, object, modifier) as a learnable embedding vector is concatenated to the token embeddings and then it is fed to the model. The embedding dimension for roles is 300.
        \item \textbf{HAQAE}~\citep{weber2018hierarchical}
         This work is the most similar to ours. For fairness, we seed HAQAE with the same dimension GRUs and pretrained embeddings. 
\end{itemize}
\subsection{Evaluations}

To measure the effectiveness of our proposed model for event representation, we first report the perplexity and Inverse Narrative Cloze metrics. %
\begin{table*}[!ht]
    \centering
    \resizebox{.98\textwidth}{!}{
    \begin{tabular}{l|ccccc}
    \specialrule{.1em}{.05em}{.05em} 
     {Tokens}& \multicolumn{5}{c}{$\beta_{\text{enc}}$ (\textbf{{Frames} Given Tokens})} \\ \cdashline{1-6}
    \textbf{kills} & \fsc{Killing} & \fsc{Death} & \fsc{Hunting\_success\_or\_failure} & \fsc{Hit\_target} & \fsc{Attack} \\ 
    \textbf{paper} & \fsc{Submitting\_documents} & \fsc{Summarizing} & \fsc{Sign} & \fsc{Deciding} & \fsc{Explaining\_the\_facts}\\
    \textbf{business} & \fsc{Commerce\_pay} & \fsc{Commerce\_buy} & \fsc{Expensiveness} & \fsc{Renting} & \fsc{Reporting} \\ 
    \specialrule{.1em}{.05em}{.05em} 
    
     {Frames} & \multicolumn{5}{c}{$\beta_{\text{dec}}$ (\textbf{Tokens Given {Frames}})} \\ \cdashline{1-6}
    \textbf{\fsc{Causation}} & raise & rendered & caused & induced & brought \\     
    \textbf{\fsc{Personal\_relationship}}& dated & dating & married & divorced & widowed \\ 
    \textbf{\fsc{Finish\_competition}}& lost & compete & won & competed & competes \\ 
    \specialrule{.1em}{.05em}{.05em}     
    \end{tabular}
    }
\caption{Results for the outputs of the attention layer, the upper table shows the $\beta_{\text{enc}}$ and the bottom table shows the $\beta_{\text{dec}}$, when $\epsilon=0.7$. Each row shows the top 5 words for each clustering. 
}
\label{tab:attentionClusters}
\end{table*} 
\paragraph{Perplexity}
We summarize our per-word perplexity results in \cref{tab:Perplexity}, which compares our event chain model, with varying $\epsilon$ values, to the three baselines.\footnote{In our case, perplexity provides an indication of the model's ability to predict the next event and arguments.} %
Recall that $\epsilon$ refers to the (average) percent of frames observed during \textit{training}. %
During evaluation \textit{no} frames are observed; this ensures a fair comparison to our baselines. %

As clearly seen, our model outperforms
other baselines across both the validation and test datasets. %
We find that increasing the observation probability $\epsilon$ consistently yields performance improvement. %
For any value of $\epsilon$ we outperform RNNLM and RNNLM+ROLE. %
HAQAE outperforms our model for $\epsilon \le 0.5$, while we outperform HAQAE for $\epsilon \in \{0.7, 0.8, 0.9\}$. %
This suggests that while the tree-based latent structure can be helpful when external, semantic knowledge is \textit{not} sufficiently available during training, a simpler linear structure can be successfully guided by that knowledge when it is available. %
Finally, recall that the external frame annotations are automatically provided, without human curation: 
this suggests that our model does not require perfectly, curated annotations. %
These observations support the hypothesis that frame observations, in conjunction with latent variables, provide a benefit to event modeling. %
\paragraph{Inverse Narrative Cloze}
This task has been proposed by \citet{weber2018hierarchical} to evaluate the ability of models to classify the legitimate sequence of events over detractor sequences. %
For this task, we have created two Wiki-based datasets from our validation and test datasets, each with 2k samples. %
Each sample has 6 options in which the first events are the same and only one of the options represents the actual sequence of events. %
All the options have a fixed size of 6 events and the one that has the lowest perplexity is selected as the correct one. %
We also consider two NYT inverse narrative cloze datasets that are publicly available.\footnote{\url{https://git.io/Jkm46}} %
All the models are trained on the Wiki dataset and then classifications are done on the NYT dataset (no NYT training data was publicly available). 

\Cref{tab:invNarrClozeStats} presents the results for this task. %
Our method tends to achieve a superior classification score over all the baselines, even for small $\epsilon$. %
Our model also yields performance improvements on the NYT validation and test datasets. We observe that the inverse narrative cloze scores for the NYT datasets is almost independent from the $\epsilon$.
We suspect this due to the different domains between training (Wikipedia) and testing (newswire).

Note that while our model's perplexity improved monotonically as $\epsilon$ increased, we do not see monotonic changes, with respect to $\epsilon$, for this task. %
By examining computed quantities from our model, we observed both that a high $\epsilon$ resulted in very low entropy attention and that frames very often attended to the verb of the event---it learned this association despite never being explicitly directly to. %
While this is a benefit to localized next word prediction (i.e., perplexity), it is detrimental to inverse narrative cloze. %
On the other hand, lower $\epsilon$ resulted in slightly higher attention entropy, suggesting that less peaky attention allows the model to capture more of the entire event sequence and improve global coherence. %

\subsection{Qualitative Analysis of Attention}
\label{sec:attention}

To illustrate the effectiveness of our proposed attention mechanism, in \cref{tab:attentionClusters} we show the most likely frames given tokens ($\beta_{\text{enc}}$) , and tokens given frames ($\beta_{\text{dec}}$). 
We define
$\beta_{\text{enc}} =
{W}_{\text{out}}\tanh(\boldsymbol{H}^{\top})$ and 
$\beta_{\text{dec}} = \hat{W}_{\text{out}}\tanh({\boldsymbol{E}^M}^{\top})$
where $\beta_{\text{enc}} \in \mathbb{R}^{F\times T}$ provides a \textit{contextual} token-to-frame soft clustering matrix for each document and analogously $\beta_{\text{dec}} \in \mathbb{R}^{V\times M}$ provides a frame-to-word soft-clustering contextualized in part based on the inferred frames.
We argue that these clusters are useful for analyzing and interpreting the model and its predictions. %
Our experiments demonstrate that the frames in the \textit{encoder} (\cref{tab:attentionClusters}, top) mostly attend to the verbs and similarly the decoder utilizes expected and reasonable frames to predict the next verb. %
Note that we have not restricted the frames and tokens connection: the attention mechanism makes the ultimate decision for these connections. 

We note that these clusters are a result of our attention mechanisms. %
Recall that in both the encoder and decoder algorithms, after computing the context vectors, we use the \textit{addition} of two $\tanh(\cdot)$ functions with the goal of separating the GRU hidden states and frame embeddings (\cref{alg:Decoder_addition}).
This is a different computation from the bi-linear attention mechanism~\citep{luong2015effective} that applies the $\tanh(\cdot)$ function over concatenation. %
Our additive approach was inspired by the neural topic modeling method from \citet{dieng2016topicrnn}, which similarly uses additive factors to learn an expressive and predictive neural component \textit{and} the classic ``topics'' (distributions/clusters over words) that traditional topic models excel at finding. %
While theoretical guarantees are beyond our scope, qualitative analyses suggests that our additive attention lets the model learn reasonable soft clusters of tokens into frame-based ``topics.'' %
See \cref{tab:AdditionBilinearPerplexity} in the Appendix for an empirical comparison and validation of our use of addition rather than concatenation in the attention mechanisms.

\subsection{How Discriminative Is A Latent Node?}
\label{sec:expt_frame_prediction}
Though we develop a generative model, we want to make sure the latent nodes are capable of leveraging the frame information in the decoder. We examine this assessing the ability of one single latent node to classify the frame for an event. %
We repurpose the Wikipedia language modeling dataset into a new training data set with 1,643,656 samples, validation with 57,261 samples and test with 75903 samples. We used 500 frame labels. Each sample is a single event. We fixed the number of latent nodes to be one. %
We use RNNLM and RNNLM+ROLE as baselines, adding a linear classifier layer followed by the softplus function on top of the bidirectional GRU last hidden vectors and a dropout of 0.15 on the logits. %
We trained all the models with the aforementioned training dataset, and tuned the hyper parameters on the validation dataset. %

\begin{table}[t]
    \centering
    \resizebox{.98\columnwidth}{!}{
  \begin{tabular}{l|c|c|c|c|c|c|c}
    \specialrule{.1em}{.05em}{.05em} 
  \multirow{2}{*}{Model} & \multirow{2}{*}{$\epsilon$} & \multicolumn{3}{c|}{Valid} & \multicolumn{3}{c}{Test} \\ \cline{3-8} 
                         &                             & Acc   & Prec  & f1 & Acc   & Prec & f1 \\ \hline
                         RNNLM & - &  0.89 
                         & 0.73 
                         & 0.66 
                         & 0.88 
                         & 0.71 
                         & 0.65 
                         \\
                         RNNLM + ROLE  & - & \textbf{0.89}
                         & 0.75 
                         & 0.69 
                         & \textbf{0.88} 
                         & 0.74 
                         & 0.68 
                         \\ \cdashline{1-8}
                         Ours & 0.00   &  0.00 
                         & 0.00 
                         & 0.00 
                         & 0.00 
                         & 0.00 
                         & 0.00 
                         \\ 
                         Ours & 0.20 & 0.59 
                         & 0.27 
                         & 0.28 
                         & 0.58 
                         & 0.27 
                         & 0.28 
                         \\ 
                         Ours & 0.40 & 0.77 
                         & 0.49  
                         & 0.50 
                         & 0.77 
                         & 0.49 
                         & 0.50 
                         \\ 
                         Ours & 0.50 & 0.79 
                         & 0.51 
                         & 0.48 
                         & 0.79 
                         & 0.50 
                         & 0.48 
                         \\ 
                         Ours & 0.70 & 0.85 
                         &0.69 
                         & 0.65 
                         & 0.84 
                         & 0.69  
                         & 0.65 
                         \\
                         Ours & 0.80 & 0.86
                         &0.77 
                         &0.74 
                         &0.85 
                         &0.76 
                         & 0.74 
                         \\
                         Ours & 0.90 & 	0.87 
                         & \textbf{0.81 }
                         & \textbf{0.78}
                         & 0.86 
                         & \textbf{0.81} 
                         & \textbf{0.79 }
                         \\                       
  \specialrule{.1em}{.05em}{.05em} 
  \end{tabular}
  }
  \caption{Accuracy and macro precision and F1-score, averaged across three different runs. 
  We present standard deviations in \cref{table:accstats}.}
  \label{table:accwostats}
\end{table}

We trained the RNNLM and RNNLM+ROLE baselines in a purely supervised way, whereas our model mixed supervised (discriminative) and unsupervised (generative) training. %
The baselines observed all of the frame labels in the training set; our model only observed frame values in training with probability $\epsilon$, which it predicted from $\bGammaM$. %
The parameters leading to the highest accuracy were chosen to evaluate the classification on the test dataset. %
The results for this task are summarized in \cref{table:accwostats}. %
Our method is an attention based model which captures all the dependencies in each event to construct the latent representation, but the baselines are autoregressive models. Our encoder acts like a discriminative classifier to predict the frames, where they will later be used in the decoder to construct the events. We expect the model performance to be comparable to RNNLM and RNNLM+ROLE in terms of classification when $\epsilon$ is high. Our model with larger $\epsilon$ tends to achieve better performance in terms of macro precision and macro F1-score. 
 \begin{figure}[t]
    \centering
    \includegraphics{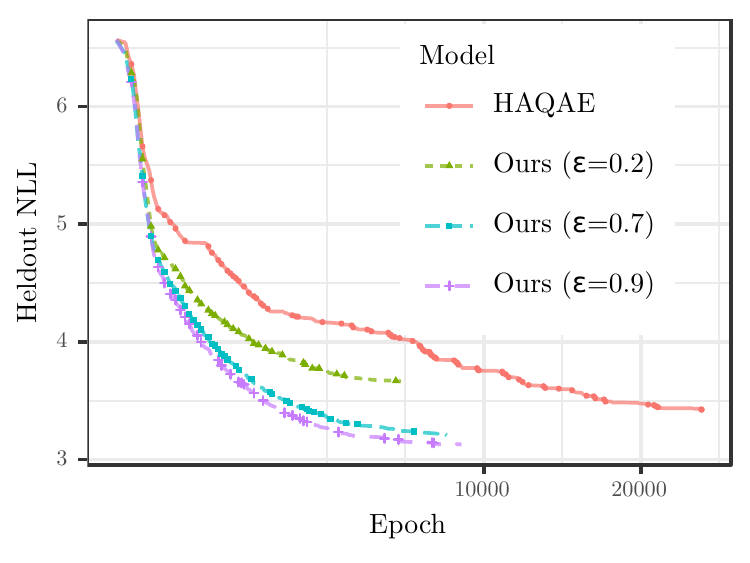}
    \caption{Validation NLL during training of our model with $\epsilon \in \{0.2,0.7,0.9\}$ and HAQAE (the red curve). Epochs are displayed with a square root transform.}
    \label{fig:nll-time}
    \vspace{-4mm}
  \end{figure}

\subsection{Training Speed}
Our experiments show that our proposed approach converges faster than the existing HAQAE model. %
For fairness, we have used the same data-loader, batch size as $100$, learning rate as $10^{-3}$ and Adam optimizer. %
In \cref{fig:nll-time}, on average each iteration takes 0.2951 seconds for HAQAE and 0.2958 seconds for our model.
From \cref{fig:nll-time} we can see that for sufficiently high values of $\epsilon$ our model is converging both better, in terms of negative log-likelihood (NLL), and faster---though for small $\epsilon$, our model still converges much faster. %
The reasons for this can be boiled down to utilizing Gumbel-Softmax rather than VQ-VAE, and also injection information in the form of frames jointly.

\section{Conclusion}
We showed how to learn a semi-supervised VAE with partially observed, sequential, discrete latent variables. We used Gumbel-Softmax and a modified attention to learn a highly effective event language model (low perplexity), predictor of how an initial event may progress (improved inverse narrative cloze), and a task-based classifier (outperforming fully supervised systems). %
We believe that future work could extend our method by incorporating other sources or types of knowledge (such as entity or ``commonsense'' knowledge), and by using other forms of a prior distribution, such as ``plug-and-play'' priors~\citep{guo2019agem,mohammadi2021ultrasound,laumont2021bayesian}.
{ 
\paragraph{Acknowledgements and Funding Disclosure}
We would also like to thank the anonymous reviewers for their comments, questions, and suggestions. %
Some experiments were conducted on the UMBC HPCF, supported by the National Science Foundation under Grant No. CNS-1920079. We'd also like to thank the reviewers for their comments and suggestions. %
This material is based in part upon work supported by the National Science Foundation under Grant Nos. IIS-1940931 and IIS-2024878. %
This material is also based on research that is in part supported by the Air Force Research Laboratory (AFRL), DARPA, for the KAIROS program under agreement number FA8750-19-2-1003. The U.S.Government is authorized to reproduce and distribute reprints for Governmental purposes notwithstanding any copyright notation thereon. The views and conclusions contained herein are those of the authors and should not be interpreted as necessarily representing the official policies or endorsements, either express or implied, of the Air Force Research Laboratory (AFRL), DARPA, or the U.S. Government.
}

\bibliography{refs}
\bibliographystyle{acl_natbib}
\clearpage

\appendix

\section{Table of Notation}
\begin{table}[h!]
\centering
\resizebox{.98\columnwidth}{!}{
\begin{tabular}{l|c|c}
 Symbol & Description & Dimension\\ \specialrule{.1em}{.05em}{.05em} 
 $F$ & Frames vocabulary size & $\mathbb{N}$ \\ \hline
 $V$ & Tokens vocabulary size& $\mathbb{N}$ \\\hline 
 $M$ &  Number of events, per sequence & $\mathbb{N}$\\ \hline
 $T$ & Number of words, per sequence $(4M)$& $\mathbb{N}$ \\ \hline
 $d_e$ & Frame emb. dim. & $\mathbb{N}$ \\ \hline
 $d_h$ & RNN hidden state dim. & $\mathbb{N}$ \\ \hline 
$\boldsymbol{E}^{F}$ & Learned frame emb. & $\mathbb{R}^{F\times d_e}$ \\ \hline
$\boldsymbol{E}^{M}$ & Embeddings of sampled frames ($\boldsymbol{E}^{M} = \mathbf{f}\boldsymbol{E}^F$) & $\mathbb{R}^{M\times d_e}$  \\ \hline
$\fval$& Observed frame (external one-hot) & $\mathbb{R}^{F}$\\ \hline
$f_m$ &  Sampled frame (simplex) & $\mathbb{R}^F$\\ \hline
$\mathbf{H}$ & RNN hidden states from the encoder & $\mathbb{R}^{T\times d_h}$\\ \hline
$z_t$ & RNN hidden state from the decoder & $\mathbb{R}^{d_h}$\\ \hline
$W_{\text{in}}$ & Learned frames-to-hidden-states weights (Encoder) &  $\mathbb{R}^{d_h\times d_e}$ \\ \hline
$\hat{W}_{\text{in}}$ & Learned hidden-states-to-frames weights (Decoder)& $\mathbb{R}^{d_e\times d_h}$ \\ \hline
$W_{\text{out}} $ & Contextualized frame emb. (Encoder) & $\mathbb{R}^{F\times d_h}$\\
\hline
$\hat{W}_{\text{out}}$ & Learned frames-to-words weights (Decoder)& $\mathbb{R}^{V\times d_e}$ \\
\hline
$\boldsymbol{\alpha}$ & Attention scores over hidden states (Encoder) & $\mathbb{R}^{T}$\\
\hline
$\boldsymbol{\hat{\alpha}}$ & Attention scores over frame emb. (Decoder) & $\mathbb{R}^{M}$\\
\hline
$\boldsymbol{c}_m$ & Context vector (Encoder)& $\mathbb{R}^{d_h}$ \\ 
\hline
$\boldsymbol{c}_t$ & Context vector (Decoder)& $\mathbb{R}^{d_e}$ \\ 
\hline
$\bGammaM$ & Gumbel-Softmax params. (Encoder) & $\mathbb{R}^{F}$\\
\hline
$w_t$ & Tokens (onehot) & $\mathbb{R}^V$\\ 
\specialrule{.1em}{.05em}{.05em} 
\end{tabular}
}
\caption{Notations used in this paper}
\label{tab:notations}
\end{table}

\section{Computing Infrastructure}
\label{app:setup_details}
We used the
Adam optimizer with initial learning rate $10^{-3}$ and early stopping (lack of validation performance improvement for 10 iterations). %
We represent events by their pretrained Glove 300 embeddings and utilized gradient clipping at 5.0 to prevent exploding gradients. %
The Gumbel-Softmax temperature is fixed to $\tau=0.5$.
We have not used dropout or batch norm on any layer. %
We have used two layers of Bi-directional GRU cells with a hidden dimension of 512 for the encoder module and Unidirectional GRU with the same configuration for the decoder. %
Each model was trained using a single GPU (a TITAN RTX RTX 2080 TI, or a Quadro 8000), though we note that neither our models nor the baselines required the full memory of any GPU (e.g., our models used roughly 6GB of GPU memory for a batch of 100 documents).

\section{Additional Insights into Novelty}
\begin{figure}[t]
    \centering
    \begin{subfigure}[b]{0.4\textwidth}
        \includegraphics[width=\textwidth]{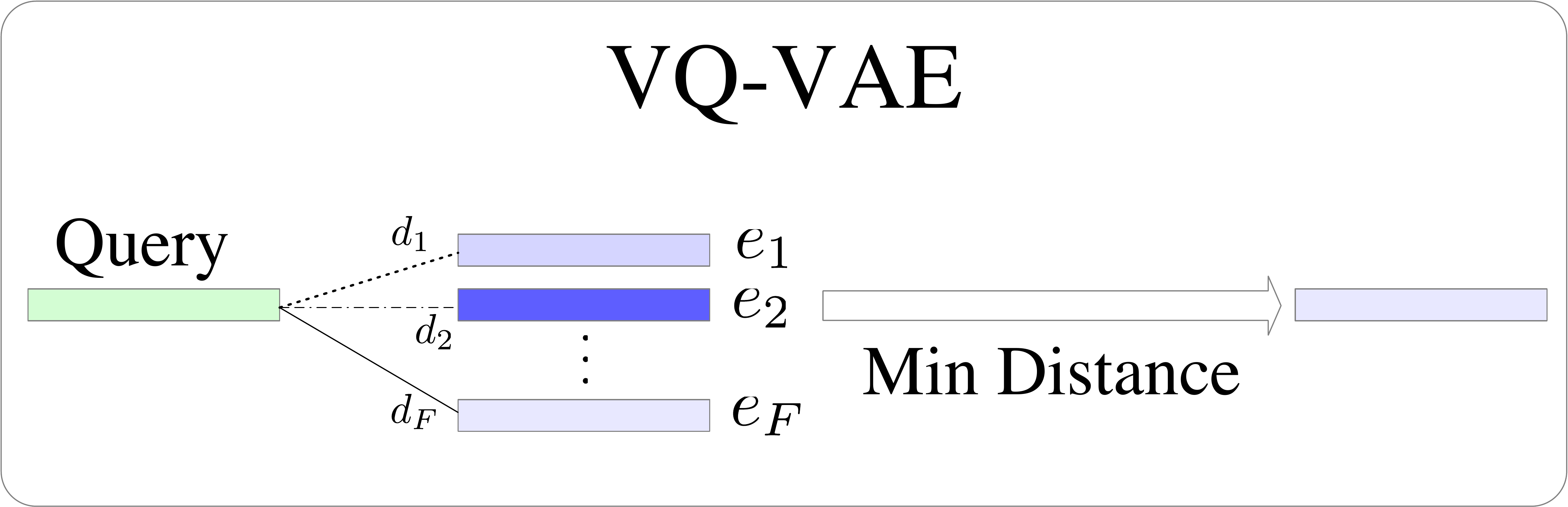}
        \caption{VQVAE}
        \label{fig:vqvae}
    \end{subfigure}
    \\[10pt] 
    \begin{subfigure}[b]{0.4\textwidth}
        \includegraphics[width=\textwidth]{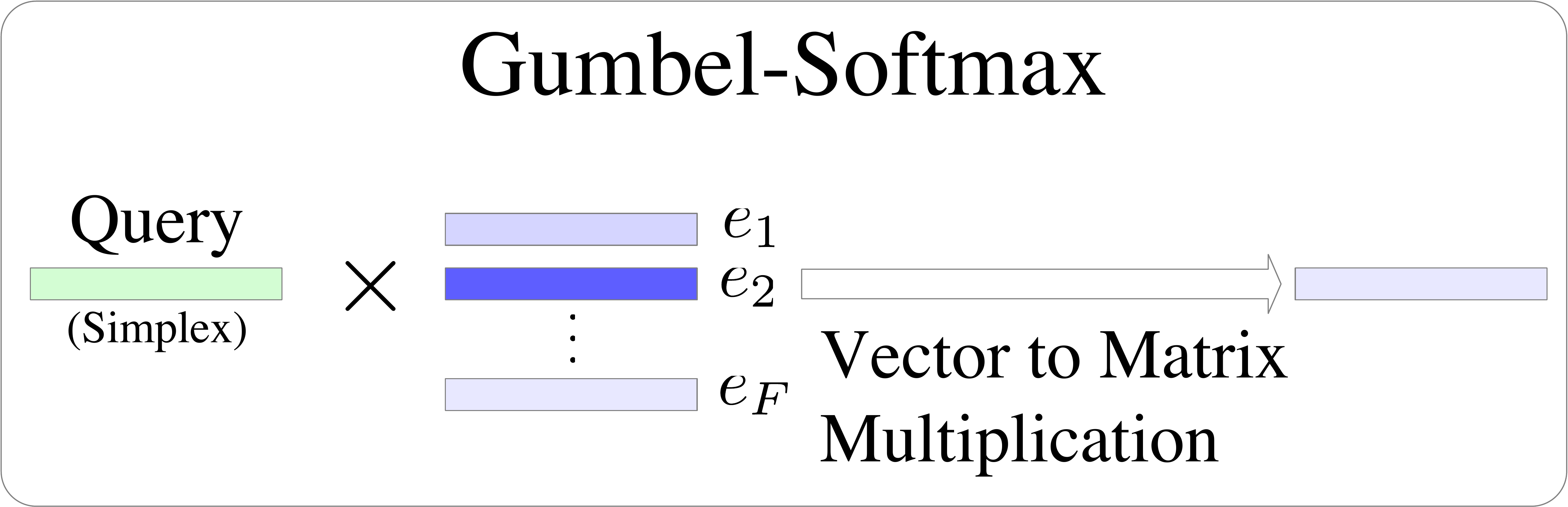}
        \caption{Gumbel-Softmax and Emb (ours)}
        \label{fig:Gumbel-softmax-emb}
    \end{subfigure}
    \caption{(a) VQ-VAE in HAQAE works based on the minimum distance between the query vector and the embeddings.
    (b) our approach first draws a Gumbel-Softmax sample and then extracts the relevant row by doing vector to matrix multiplication.
    }
    \label{fig:haqae-and-Gumbel-softmax-emb}
  \end{figure}

We have previously mentioned how our work is most similar to HAQAE~\citep{weber2018hierarchical}. In this section, we provide a brief overview of HAQAE (\cref{app:haqae_review}) and then highlight differences (\cref{fig:haqae-and-Gumbel-softmax-emb}), with examples (\cref{app:frame_vector_norm}).

\subsection{Overview of HAQAE}
\label{app:haqae_review}
HAQAE provides an unsupervised tree structure based on the vector quantization variational autoencoder (VQVAE) over $M$ latent variables. Each latent variable $z_i$ is defined in the embedding space denoted as $e$. The varaitional distribution to approximate $\textbf{z}=\{z_1,z_2,\ldots,z_M\}_{i=1}^{M}$ given tokens $x$ is defined as follows:
\[
q(\textbf{z} | x) = q_0(z_0 | x) \prod^{M-1}_{i=1} q_i(z_i | \text{parent\_of}(z_{i}), x).
\]
The encoder calculates the attention over the input RNN hidden states $\textbf{h}_x$ and the parent of $z_{i}$ to define  
$q_i(z_i=k|z_{i-1}, x)$:
\[
\begin{cases}
1 & \text{k} = \text{argmin}_j \|g_i(x, \text{parent\_of}(z_{i})) - e_{ij}\|_2 \\
0 & \text{elsewise},
\end{cases}
\]
where $g_i(x, \text{parent\_of}(z_{i}))$ computes bilinear attention between $\textbf{h}_x$ and $\text{parent\_of}(z_{i}))$.

In this setting, the variational distribution is deterministic; right after deriving the latent query vector it will be compared with a lookup embedding table and the row with minimum distance is selected, see \cref{fig:vqvae}. The decoder reconstructs the tokens as $p(x_i\vert \textbf{z})$ 
that calculates the attention over latent variables and the RNN hidden states.
A reconstruction loss $L^R_j$ and a ``commit'' loss~\citep{weber2018hierarchical} loss $L^C_j$ force $g_j(x, \text{parent\_of}(z_{i}))$ to be close to the embedding referred to by $q_i(z_i)$. %
\begin{table*}
    \centering
    \resizebox{.98\textwidth}{!}{
  \begin{tabular}{l|c|c|c|c|c|c|c}
    \specialrule{.1em}{.05em}{.05em} 
  \multirow{2}{*}{Model} & \multirow{2}{*}{$\epsilon$} & \multicolumn{3}{c|}{Valid} & \multicolumn{3}{c}{Test} \\ \cline{3-8} 
                         &                             & Acc   & Prec  & f1 & Acc   & Prec & f1 \\ \hline
                         RNNLM & - &  0.89  $\pm{0.001 }$
                         & 0.73  $\pm{0.004 }$
                         & 0.66  $\pm{0.008}$
                         & 0.88  $\pm{0.005}$
                         & 0.71  $\pm{0.014}$
                         & 0.65  $\pm{0.006}$
                         \\
                         RNNLM + ROLE  & - & 0.89 $\pm{0.004 }$
                         & 0.75  $\pm{0.022}$
                         & 0.69  $\pm{0.026}$
                         & 0.88  $\pm{0.005}$
                         & 0.74  $\pm{0.017}$
                         & 0.68  $\pm{0.020}$
                         \\ \cdashline{1-8}
                         Ours & 0.00   &  0.00  $\pm{0.001 }$
                         & 0.00  $\pm{0.001}$
                         & 0.00  $\pm{0.000}$
                         & 0.00  $\pm{0.001}$
                         & 0.00  $\pm{0.000}$
                         & 0.00  $\pm{0.000}$
                         \\ 
                         Ours & 0.20 & 0.59  $\pm{0.020  }$
                         & 0.27  $\pm{0.005}$
                         & 0.28  $\pm{0.090}$
                         & 0.58  $\pm{0.010}$
                         & 0.27  $\pm{0.050}$
                         & 0.28  $\pm{0.010}$
                         \\ 
                         Ours & 0.40 & 0.77  $\pm{0.012  }$
                         & 0.49   $\pm{0.130  }$
                         & 0.50  $\pm{0.010  }$
                         & 0.77  $\pm{0.010  }$
                         & 0.49  $\pm{0.060   }$
                         & 0.50  $\pm{0.010   }$
                         \\ 
                         Ours & 0.50 & 0.79  $\pm{0.010  }$
                         & 0.51  $\pm{0.080  }$
                         & 0.48  $\pm{0.080  }$
                         & 0.79  $\pm{0.009  }$
                         & 0.50  $\pm{0.080   }$
                         & 0.48  $\pm{0.050  }$
                         \\ 
                         Ours & 0.70 & 0.85  $\pm{0.007 }$
                         &0.69  $\pm{0.050  }$
                         & 0.65  $\pm{0.040 }$
                         & 0.84  $\pm{0.013 }$
                         & 0.69   $\pm{0.050  }$
                         & 0.65  $\pm{0.020 }$
                         \\
                         Ours & 0.80 & 0.86 $\pm{0.003 }$
                         &0.77  $\pm{0.013   }$
                         &0.74  $\pm{0.006  }$
                         &0.85  $\pm{0.005}$
                         &0.76  $\pm{0.008  }$
                         & 0.74  $\pm{0.010}$ \\
                         
                         Ours & 0.90 & 	0.87  $\pm{0.002 }$
                         & 0.81  $\pm{0.007   }$
                         & 0.78 $\pm{0.006}$
                         & 0.86  $\pm{0.011 }$
                         & 0.81  $\pm{0.020  }$
                         & 0.79  $\pm{0.010  }$
                         \\                       
  \specialrule{.1em}{.05em}{.05em} 
  \end{tabular}
  }
  \caption{Accuracy and macro precision and F1-score, averaged across 3 runs, with standard deviation reported.}
  \label{table:accstats}
\end{table*}
Both $L^R_j$ and $L^C_j$ terms rely on a deterministic mapping that is based on a nearest neighbor computation that makes it difficult to inject guiding information to the latent variable.

\subsection{Frame Vector Norm}
\label{app:frame_vector_norm}
Like HAQAE, we use embeddings $\boldsymbol{E}^{F}$ instead of directly using the Gumbel-Softmax frame samples.
In our model definition, each frame $f_m=[f_{m,1},f_{m,2},\ldots,f_{m,F}]$ sampled from the 
Gumbel-Softmax distribution
is a simplex vector:
\begin{equation}
0\le f_{m,i}\le1,\quad \sum_{i=1}^{F}{f_{m,i}}=1.
\end{equation}
So 
$
\|
f_m
\|
_p
=
{
\big(
\sum_{i=1}^{F}
f_{m,i}^p
\big)
}
^
{
1/p
}
\leq 
F
^
{1/p}.
$
After sampling a frame simplex vector, we multiply it to the frame embeddings matrix $\boldsymbol{E}^{F}$. With an appropriate temperature for Gumbel-Softmax, the simplex would be approximately a one-hot vector and  
the multiplication maps the simplex vector to the embeddings space without any limitation on the norm.

\begin{table}[t]
    \centering
    \small
    \resizebox{\columnwidth}{!}{
\begin{tabular}{l|l|l|l|l}
\specialrule{.1em}{.1em}{0.005em}
\multicolumn{1}{c|}{\multirow{2}{*}{$\epsilon$}} & \multicolumn{2}{c|}{Addition}                          & \multicolumn{2}{c}{Concatenation}                     \\ \cline{2-5} 
\multicolumn{1}{c|}{}                     & \multicolumn{1}{c|}{Valid} & \multicolumn{1}{c|}{Test} & \multicolumn{1}{c|}{Valid} & \multicolumn{1}{c}{Test} \\ \hline
0.0 &    41.18 $\pm{ 0.69 }$    &  36.28 $\pm{ 0.74 }$      &   48.86 $\pm{0.13}$     & 42.76 $\pm{0.08}$        \\ 
0.2 &   38.52 $\pm{ 0.83 }$      &  33.31 $\pm{ 0.63}$       &   48.73 $\pm{0.18}$     & 42.76 $\pm{0.06}$        \\ 
0.4 &   37.79 $\pm{0.52}$        & 33.12 $\pm{0.54}$ &   49.34 $\pm{0.32}$      & 43.18 $\pm{0.21}$        \\ 
0.5 &  35.84 $\pm{0.66}$        &  31.11 $\pm{0.85}$  &   49.83 $\pm{0.24}$      & 43.90 $\pm{0.32}$        \\ 
0.7 &  24.20 $\pm{1.07 }$       &  21.19 $\pm{0.76}$       &   52.41 $\pm{0.38}$       & 45.97 $\pm{0.48}$        \\
0.8 &  23.68 $\pm{0.75}$        &  20.77 $\pm{0.73}$  &   55.13 $\pm{0.31}$       & 48.27 $\pm{0.21}$        \\   
0.9 &   {22.52}$\pm{{0.62}}$  & 19.84 $\pm{{0.52}}$ &   58.63 $\pm{0.25}$       & 51.39 $\pm{0.34}$        \\
\specialrule{.1em}{.1em}{0.005em}
\end{tabular}
}
    \caption{Validation and test per-word perplexities with bilinear-attention (lower is better).  %
    }
    \label{tab:AdditionBilinearPerplexity}
\end{table}

\section{More Results}

In this section, we provide additional quantitative and qualitative results, to supplement what was presented in \cref{sec:Experimental Result}.

\subsection{Standard Deviation for Frame Prediction (\cref{sec:expt_frame_prediction})}

In \cref{table:accstats}, we see the average results of classification metrics with their corresponding standard deviations. 

\subsection{Importance of Using Addition rather than Concatenation in Attention}

To demonstrate the effectiveness of our proposed attention mechanism, we compare the addition against the concatenation (regular bilinear attention) method. %
We report the results on the Wikipedia dataset in \cref{tab:AdditionBilinearPerplexity}. %
Experiments indicate that the regular bilinear attention structure with larger $\epsilon$ obtains worse performance. %
These results confirm the claim that the proposed approach benefits from the addition structure.

\subsection{Generated Sentences}
Recall that the reconstruction loss is
\begin{align*}
    \mathcal{L}_w &
    \approx    
    \dfrac{1}{S}
    \sum_{s}
    \log
    {
        p(\bW\vert f_1^{(s)},f_2^{(s)},\ldots f_M^{(s)}; z)
    },
\end{align*}
where $f_{m}^{(s)}\sim q(f_m\vert f_{m-1}^{(s)}, \fval, \bW)$. Based on these formulations,
we provide some examples of generated scripts. %
Given the seed, the model first predicts the first frame $f_1$, then it predicts the next verb $v_1$  and similarly samples the tokens one-by-one. %
During the event generations, if the sampled token is unknown, the decoder samples again. As we see in \cref{tab:gen_scripts}, the generated events and frames samples are consistent which shows the ability of model in event representation.

\begin{table*}
    \centering
    \resizebox{.99\textwidth}{!}{    
    \begin{tabular}{l|l|l|l}
    \specialrule{.1em}{.05em}{.05em} 
    \multicolumn{2}{c|}{\textbf{Seed}} & \multicolumn{1}{l|}{Event 1} & \multicolumn{1}{l}{Event 2} \\ \hline
    \multicolumn{2}{l|}{\textbf{elected taylor election at} [\fsc{Change\_of\_leadership}]} & \multicolumn{1}{l|}{served she attorney as [\fsc{Assistance}]} & \multicolumn{1}{l}{sworn she house to [\fsc{Commitment}]} \\   
    \multicolumn{2}{l|}{\textbf{graduated ford berkeley at} [\fsc{Scrutiny}]} & earned he theology in [\fsc{Earnings\_and\_losses}] & documented he book in [\fsc{Recording}] \\
    \multicolumn{2}{l|}{\textbf{released album 2008 in} [\fsc{Releasing}]} & \multicolumn{1}{l|}{helped song lyrics none [\fsc{Assistance}]} & \multicolumn{1}{l}{made chart song with [\fsc{Causation}]} \\ 
    \multicolumn{2}{l|}{\textbf{created county named and} [\fsc{Intentionally\_create}]} & \multicolumn{1}{l|}{totals district population of [\fsc{Amounting\_to}]} & \multicolumn{1}{l}{served district city none [\fsc{Assistance}]} \\ 
    \multicolumn{2}{l|}{\textbf{published he december on} [\fsc{Summarizing}]} & \multicolumn{1}{l|}{written book published and [\fsc{Text\_creation}]} & \multicolumn{1}{l}{place book stories among [\fsc{Placing}]} \\ 
    \multicolumn{2}{l|}{\textbf{played music show on} [\fsc{Performers\_and\_roles}]} & starred film music none [\fsc{Performers\_and\_roles}] & tend lyrics music as [\fsc{Likelihood}] \\ 
    \multicolumn{2}{l|}{\textbf{aired series canada in} [\fsc{Expressing\_publicly}]} & features series stories none [\fsc{Inclusion}] & abused characters film in [\fsc{Abusing}] \\ 
    \multicolumn{2}{l|}{\textbf{consists band members of} [\fsc{Inclusion}]} & belong music party to [\fsc{Membership}] & leads music genre into [\fsc{Causation}] \\ 
    \specialrule{.1em}{.05em}{.05em}     
    \end{tabular}
    }
    \caption{Generated scripts and the inferred frames (in brackets) from the seed event in boldface ($\epsilon=0.7$)}
    \label{tab:gen_scripts}
    \end{table*}

\subsection{Inferred Frames}
Using \cref{alg:Encoder}, we can see the frames sampled during the training and validation.
In \cref{tab:infer_frames}, we provide some examples of frame inferring for both training and validation examples. We observe that for training examples when $\epsilon>0.5$, almost all the observed frames and inferred frames are the same. In other words, the model prediction is almost the same as the ground truth.

Interestingly, the model is more flexible in sampling the latent frames (orange ones). In \cref{tab:infer_ex1}
the model is estimating \fsc{have\_associated} instead of the \fsc{possession} frame.
 In \cref{tab:infer_ex2}, instead of \fsc{process\_start} we have \fsc{activity\_start} and finally in
 \cref{tab:infer_ex4} we have \fsc{taking\_sides} rather than \fsc{supporting}.
 
Some wrong predictions like \fsc{catastrophe} instead of \fsc{causation} in \cref{tab:infer_ex3} can be considered as the effect of other words like ``\textit{pressure}" in ``\textit{resulted pressure revolution in}". In some cases like \cref{tab:infer_ex2} the predicted frame \fsc{being\_named} is a better choice than the ground truth \fsc{appointing}. In the same vein \fsc{finish\_competition} is a better option rather than \fsc{getting} in \cref{tab:infer_ex6}.

\begin{table}[t]
    \begin{subtable}[t]{.5\textwidth}
    \centering
    \scalebox{0.43}{  
    \begin{tabular}{l|l|l|l}
    \specialrule{.1em}{.05em}{.05em} 
    \multicolumn{2}{c|}{Event} & Ground Truth& Inferred Frame \\ \hline
    1 & estimated product prices in & \fsc{\blue{Estimating}} & \fsc{{Estimating}} \\ \hline
    2 & rose which billions to  & \fsc{\blue{Change\_position\_on\_a\_scale}} & \fsc{{Change\_position\_on\_a\_scale}} \\ \hline
    3 & had maharashtra per as\_of  & \fsc{\blue{Possession}} & \fsc{{Possession}} \\ \hline
    4 & houses mumbai headquarters none  & \fsc{\blue{Buildings}} & \fsc{{Buildings}} \\ \hline
    5 & have \% offices none & \fsc{\orange{Possession}} & \fsc{{Have\_associated}} \\ 
    \specialrule{.1em}{.05em}{.05em} 
    \end{tabular}
    }
    \caption{Example 1 (training)}
    \label{tab:infer_ex1}
    \end{subtable}
    \\[4pt]
    \begin{subtable}[t]{.5\textwidth}
    \centering
    \scalebox{0.65}{  
    \begin{tabular}{l|l|l|l}
    \specialrule{.1em}{.05em}{.05em} 
    \multicolumn{2}{c|}{Event} & Ground Truth& Inferred Frame \\ \hline
    1 & competed she olympics in & \fsc{\blue{Win\_prize}} & \fsc{{Win\_prize}} \\ \hline
    2 & set she championships at & \fsc{\blue{Cause\_to\_start}} & \fsc{{Cause\_to\_start}} \\ \hline
    3 & named she <unk> in & \fsc{\orange{Appointing}} & \fsc{{Being\_named}} \\ \hline
    4 & started she had but & \fsc{\orange{Process\_start}} & \fsc{{Activity\_start}} \\ \hline
    5 & had she height because\_of & \fsc{\blue{Possession}}  & \fsc{{Possession}} \\ 
    \specialrule{.1em}{.05em}{.05em} 
    \end{tabular}
    }
    \caption{Example 2 (training)}
    \label{tab:infer_ex2}
    \end{subtable}
    \\[4pt]
    \begin{subtable}[t]{.5\textwidth}
    \centering
    \scalebox{0.61}{  
    \begin{tabular}{l|l|l|l}
    \specialrule{.1em}{.05em}{.05em} 
    \multicolumn{2}{c|}{Event} & Ground Truth& Inferred Frame \\ \hline
    1 & arose that revolution after & \fsc{\orange{Coming\_to\_be}} & \fsc{{Coming\_to\_be}} \\ \hline
    2 & supported who charter none & \fsc{\blue{Taking\_sides}} & \fsc{{Taking\_sides}} \\ \hline
    3 & developed pressure classes from & \fsc{\blue{Progress}} & \fsc{{Progress}} \\ \hline
    4 & remained monarchy run and & \fsc{\orange{State\_continue}} & \fsc{{State\_continue}} \\ \hline
    5 & resulted pressure revolution in & \fsc{\orange{Causation}} & \fsc{{Catastrophe}} \\ 
    \specialrule{.1em}{.05em}{.05em} 
    \end{tabular}
    }
    \caption{Example 3 (training)}
    \label{tab:infer_ex3}
    \end{subtable}
    \\[4pt]
    \begin{subtable}[t]{.5\textwidth}
    \centering
    \scalebox{0.64}{  
    \begin{tabular}{l|l|l|l}
    \specialrule{.1em}{.05em}{.05em} 
    \multicolumn{2}{c|}{Event} & Ground Truth& Inferred Frame \\ \hline
    1 & features it form in & \fsc{\orange{Inclusion}} & \fsc{{Inclusion}} \\ \hline
    2 & allows format provides and & \fsc{\orange{Permitting}} & \fsc{{Deny\_permission}} \\ \hline
    3 & provides format forum none & \fsc{\orange{Supply}} & \fsc{{Supply}} \\ \hline
    4 & rely readers staff upon & \fsc{\orange{Reliance}} & \fsc{{Reliance}} \\ \hline
    5 & support citations assertions none & \fsc{\orange{Supporting}} & \fsc{{Taking\_sides}} \\
    \specialrule{.1em}{.05em}{.05em} 
    \end{tabular}
    }
    \caption{Example 4 (validation)}
    \label{tab:infer_ex4}
    \end{subtable}
    \\[4pt]
    \begin{subtable}[t]{.5\textwidth}
    \centering
    \scalebox{0.52}{  
    \begin{tabular}{l|l|l|l}
    \specialrule{.1em}{.05em}{.05em} 
    \multicolumn{2}{c|}{Event} & Ground Truth& Inferred Frame \\ \hline
    1 & attended he university none & \fsc{\orange{Attending}} & \fsc{{Attending}} \\ \hline
    2 & moved he 1946 in & \fsc{\orange{Motion}} & \fsc{{Travel}} \\ \hline
    3 & married he yan in & \fsc{\orange{Personal\_relationship}} & \fsc{{Forming\_relationships}} \\ \hline
    4 & worked they moved and & \fsc{\orange{Being\_employed}} & \fsc{{Being\_employed}} \\ \hline
    5 & moved they beijing to & \fsc{\orange{Motion}} & \fsc{{Motion}} \\
    \specialrule{.1em}{.05em}{.05em} 
    \end{tabular}
    }
    \caption{Example 5 (validation)}
    \label{tab:infer_ex5}
    \end{subtable}
    \\[4pt]
    \begin{subtable}[t]{.5\textwidth}
    \centering
    \scalebox{0.52}{  
    \begin{tabular}{l|l|l|l}
    \specialrule{.1em}{.05em}{.05em} 
    \multicolumn{2}{c|}{Event} & Ground Truth& Inferred Frame \\ \hline
    1 & had he achievements none & \fsc{\orange{Possession}} & \fsc{{Have\_associated}} \\ \hline
    2 & competed he finished and & \fsc{\orange{Finish\_competition}} & \fsc{{Required\_event}} \\ \hline
    3 & finished he relay in & \fsc{\orange{Process\_completed\_state}} & \fsc{{Activity\_done\_state}} \\ \hline
    4 & set he three between & \fsc{\orange{Intentionally\_create}} & \fsc{{Intentionally\_create}} \\ \hline
    5 & won he championships at & \fsc{\orange{Getting}} & \fsc{{Finish\_competition}} \\
    \specialrule{.1em}{.05em}{.05em} 
    \end{tabular}
    }
    \caption{Example 6 (validation)}
    \label{tab:infer_ex6}
    \end{subtable}
    \caption{Sequences of 5 events and the inferred frames during training with partial frame observation, and validation without any observation. Blue frames are observed and orange frames are latent.}
    \label{tab:infer_frames}
    \end{table}

\subsection{Clustering}

\begin{table*}
    \centering
    \rotatebox{90}{
    \resizebox{7.75in}{!}{ 
    \begin{tabular}{l|ccccc}
    \specialrule{.1em}{.05em}{.05em} 
     {Verbs}& \multicolumn{5}{c}{$\beta_{\text{enc}}$(\textbf{\fsc{Frames} Given Verbs})} \\[6pt] \cdashline{1-6}
    \textbf{executed} & \fsc{Execution} & \fsc{Verdict} & \fsc{Hiring} & \fsc{Rite} & \fsc{Being\_operational} \\[3pt]
    \textbf{start} & \fsc{Activity\_start} & \fsc{Process\_start} & \fsc{Duration\_relation} & \fsc{Activity\_done\_state} & \fsc{Relative\_time} \\[3pt] 
    \textbf{fought} & \fsc{Hostile\_encounter} & \fsc{Participation} & \fsc{Success\_or\_failure} & \fsc{Quitting} & \fsc{Death} \\[3pt] 
   \textbf{worked} & \fsc{Being\_employed} & \fsc{Working\_on} & \fsc{Hiring} & \fsc{Being\_operational} & \fsc{Education\_teaching} \\[3pt] 
    \textbf{made} & \fsc{Causation} & \fsc{Manufacturing} & \fsc{Intentionally\_create} & \fsc{Cause\_change} & \fsc{Creating} \\[3pt] 
    \textbf{dominated} & \fsc{Dominate\_situation} & \fsc{Control} & \fsc{Boundary} & \fsc{Separating} & \fsc{Invading} \\[3pt]
    \textbf{created} & \fsc{Intentionally\_create} & \fsc{Creating} & \fsc{Cause\_change} & \fsc{Coming\_to\_be} & \fsc{Causation} \\[3pt]
    \textbf{houses} & \fsc{Buildings} & \fsc{Locative\_relation} & \fsc{Residence} & \fsc{Commerce\_buy} & \fsc{Cause\_expansion} \\[3pt]
    \textbf{see} & \fsc{Perception\_experience} & \fsc{Grasp} & \fsc{Evidence} & \fsc{Similarity} & \fsc{Experiencer\_obj} \\[3pt]
    \textbf{start} & \fsc{Activity\_start} & Process\_start & Secrecy\_status & Sign\_agreement & Activity\_finish \\[3pt]
    \textbf{gave} & \fsc{Giving} & \fsc{Getting} & \fsc{Bringing} & \fsc{Supply} & \fsc{Needing} \\[3pt]
    \textbf{passed} & \fsc{Traversing} & \fsc{Undergo\_change} & \fsc{Control} & \fsc{Change\_of\_leadership} & \fsc{Arriving} \\[3pt]
    \textbf{sold} & \fsc{Commerce\_sell} & \fsc{Commerce\_buy} & \fsc{Manufacturing} & \fsc{Prevent\_from\_having} & \fsc{Releasing} \\[3pt]
    \textbf{rose} & \fsc{Motion\_directional} & \fsc{Change\_position\_on\_a\_scale} & \fsc{Getting\_up} & \fsc{Boundary} & \fsc{Expansion}\\[3pt]
    \textbf{returned} & \fsc{Arriving} & \fsc{Traversing} & \fsc{Intentionally\_affect} & \fsc{Have\_associated} & \fsc{Getting} \\[3pt]
    \textbf{enlisted} & \fsc{Becoming\_a\_member} & \fsc{Collaboration} & \fsc{Appointing} & \fsc{Cotheme} & \fsc{Come\_together} \\[3pt]
    \textbf{met} & \fsc{Come\_together} & \fsc{Meet\_with} & \fsc{Make\_acquaintance} & \fsc{Discussion} & \fsc{Boundary} \\[3pt]
    \textbf{entered} & \fsc{Activity\_start} & \fsc{Arriving} & \fsc{Process\_start} & \fsc{Undergo\_change} & \fsc{Mass\_motion} \\[3pt]
    \textbf{obtained} & \fsc{Getting} & \fsc{Commerce\_buy} & \fsc{Giving} & \fsc{Conquering} & \fsc{Earnings\_and\_losses} \\[3pt]
    \textbf{chosen} & \fsc{Choosing} & \fsc{Volubility} & \fsc{Leadership} & \fsc{Be\_subset\_of} & \fsc{Coming\_up\_with} \\[10pt]

    \specialrule{.1em}{.05em}{.05em} 
     {Frames} & \multicolumn{5}{c}{$\beta_{\text{dec}}$(\textbf{Tokens Given \fsc{Frames}})} \\[6pt] \cdashline{1-6}
    \textbf{\fsc{Becoming}} & become & becomes & became & turned & goes  \\[3pt] 
    \textbf{\fsc{Shoot\_projectiles}}& launched & fired & shoots & shoot & rocket \\[3pt] 
    \textbf{\fsc{Judgment\_communication}}& criticized& charged & criticised & praised & accused \\[3pt]
    \textbf{\fsc{Shoot\_projectiles}}& launched & fired & shoots & shoot & rocket \\[3pt] 
    \textbf{\fsc{Judgment\_communication}}& criticized& charged & criticised & praised & accused \\[3pt]    
    \textbf{\fsc{Being\_named}} & named & entitled & known & called & honour \\[3pt]
    \textbf{\fsc{Assistance}} & serve & serves & served & helps & assisted \\[3pt]
    \textbf{\fsc{Change\_of\_leadership}} & elected & elects & councillors & installed & overthrew \\[3pt]
    \textbf{\fsc{Reveal\_secret}} & exposed & revealed & disclosed & reveal & confessed \\[3pt]
    \textbf{\fsc{Text\_creation}} & composed & authored & composes & drafted & written \\[3pt]
    \textbf{\fsc{Expressing\_publicly}} & aired & voiced & expresses & express & airs \\[3pt]
    \textbf{\fsc{Arrest}} & arrested & apprehended & booked & jailed & re-arrested \\[3pt]
    \textbf{\fsc{Collaboration}} & cooperate & collaborated & teamed & partnered & cooperated \\[3pt]
    \textbf{\fsc{Win\_prize}} & competing & compete & competes & competed & mid-1950s \\[3pt]
    \textbf{\fsc{Leadership}} & governs & administered & administers & presides & ruled \\[3pt]
    \textbf{\fsc{Inclusion}} & contains & included & includes & include & excluded \\[3pt]
    \textbf{\fsc{Self\_motion}} & sailed & walks & ran & danced & walk \\[3pt]
    \textbf{\fsc{Ingestion}} & eaten & drank & drink & consumed & eat \\[3pt]
    \textbf{\fsc{Taking\_sides}} & supported & sided & supporting & endorsed & endorses \\[3pt]
    \specialrule{.1em}{.05em}{.05em}     
    \end{tabular}
    }
    }
\caption{Results for the outputs of the attention layer, the upper table shows the $\beta_{\text{enc}}$ and the bottom table shows the $\beta_{\text{dec}}$, when $\epsilon=0.7$. Each row shows the top 5 words for each clustering. }
\label{tab:moreClusters}
\end{table*} 

Here we provide more examples for $\beta_{\text{enc}}$ and $\beta_{\text{dec}}$ in \cref{tab:moreClusters}. Our experiments show that by sorting the tokens in each frame, the first 20 words are mostly verbs. And among different token types, verbs are better classifiers for frames.

\end{document}